\documentclass{article}

\usepackage[eandd,preprint]{neurips_2026}
\usepackage[utf8]{inputenc}
\usepackage[T1]{fontenc}
\usepackage{hyperref}
\usepackage{url}
\usepackage{amsfonts}
\usepackage{nicefrac}
\usepackage{microtype}
\usepackage[table]{xcolor}
\usepackage{graphicx}
\usepackage{multirow}
\usepackage{enumitem}
\usepackage{adjustbox}
\usepackage{makecell}
\usepackage{subcaption}
\usepackage{booktabs}
\usepackage{tabularx}
\usepackage{array}
\usepackage{marvosym} 

\definecolor{bestcolor}{HTML}{1a5276}

\title{EgoMonth: A Month-Level Egocentric Video Benchmark for Long-Term Spatiotemporal Memory}

\author{
Weitao Chen\textsuperscript{1,*},
Hu Jiaxin\textsuperscript{1,*},
Xie Tianyidan\textsuperscript{1},
Yang Li\textsuperscript{1},
Yuyi Qian\textsuperscript{1},
Banghao Xu\textsuperscript{1},\\
Ziheng Tang\textsuperscript{1},
Shenyi Wang\textsuperscript{1},
Mingyue Yu\textsuperscript{1},
Duo Li\textsuperscript{1},
Jiacheng Shi\textsuperscript{1},
Gao Wang\textsuperscript{1},\\
Zhan Xu\textsuperscript{2},
Zhicheng Qiu\textsuperscript{2},
Xuanfu Li\textsuperscript{2},
Jian Yang\textsuperscript{1},
Lanjun Wang\textsuperscript{3,\Letter},
Zili Yi\textsuperscript{1,\Letter}
\\[2mm]
\textsuperscript{1}Nanjing University, China
\qquad
\textsuperscript{2}Huawei Technologies Co., Ltd., China
\\
\textsuperscript{3}Tianjin University, China
\\[1mm]
\texttt{522025710006@smail.nju.edu.cn}
\qquad
\texttt{yi@nju.edu.cn}
\\[2mm]
\textsuperscript{*}Equal contribution.
\qquad
\Letter\ Corresponding authors.
}

\begin{document}

\maketitle

\begin{figure}[ht]
  \centering
  \includegraphics[width=\linewidth]{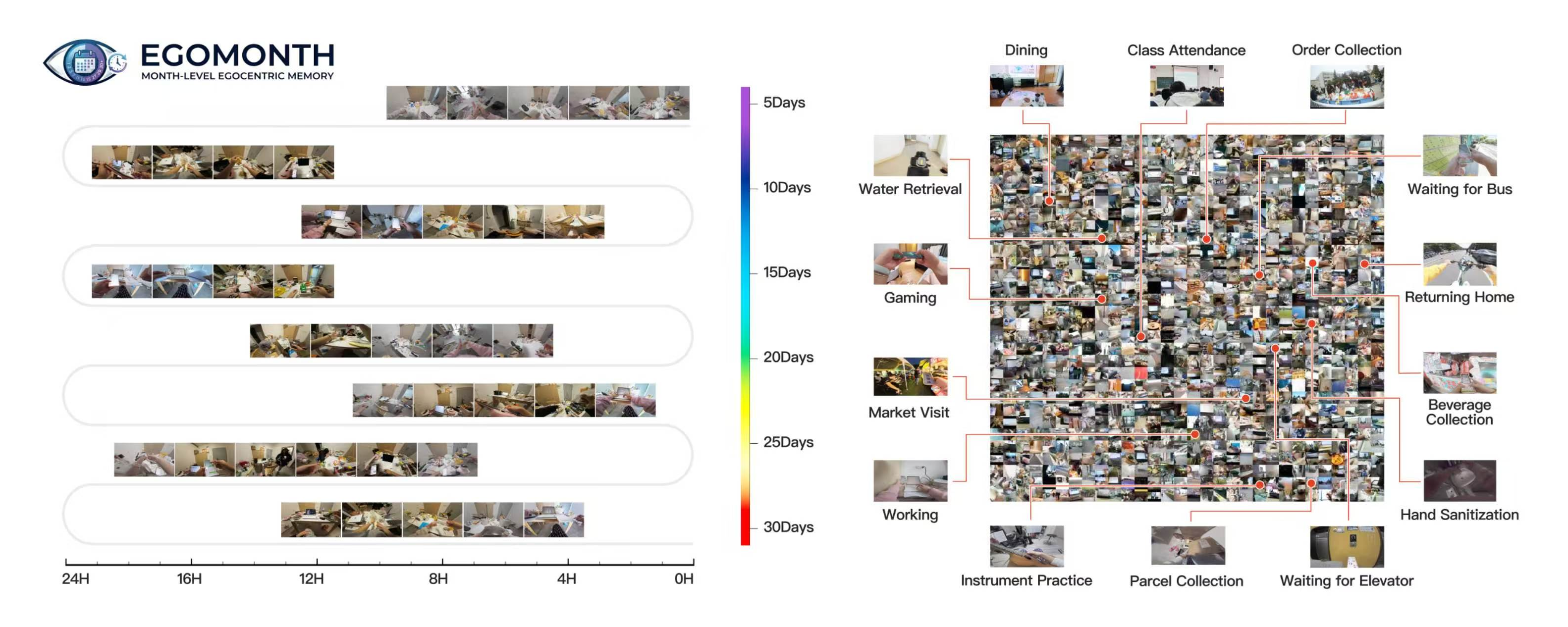} 
  \caption{\textbf{Overview of the EgoMonth dataset.} 
EgoMonth collects month-level egocentric videos from real-world daily life to evaluate long-term spatiotemporal memory.
The left panel highlights month-scale spatiotemporal continuity across daily first-person recordings, while the right panel illustrates the richness of everyday scenes, activities, and events captured in the benchmark.}
  \label{fig:teaser}
\end{figure}

\begin{abstract}
Recent advances in Multimodal Large Language Models (MLLMs) have led to substantial progress in video understanding, accompanied by a growing number of long video benchmarks. However, existing benchmarks rely predominantly on web-sourced videos that lack inter-clip spatiotemporal continuity, making it difficult to assess whether models can maintain consistent memory across days or weeks of real-world experience. We introduce \textbf{EgoMonth}, the first month-level egocentric video understanding benchmark. EgoMonth comprises over 300 hours of first-person daily-life recordings from 20 participants spanning 20 to 120 days, paired with 1{,}443 human-crafted multiple-choice QA pairs. We design a cognitively grounded \textbf{14-task} evaluation framework organized into three hierarchical cognitive levels: Schema Consolidation, Episodic Indexing, and Cascading Reasoning. Evaluation of state-of-the-art open-source and closed-source MLLMs reveals that even the best-performing model, Gemini 2.5 Pro, achieves only 71.8\% macro-average accuracy, remaining 22.4 percentage points below the human baseline of 94.2\%. Several models perform near or below the 25\% chance level on tasks such as \emph{Route Reasoning}, \emph{Cross-view Spatial Reasoning}, and \emph{Direction Judgement}, while even the strongest evaluated model remains substantially below human performance. These results indicate that current MLLMs function as \emph{lossy summarizers} rather than \emph{faithful memorizers}, highlighting the need for architectures with genuine long-term spatiotemporal memory.
\end{abstract}


\section{Introduction}
\label{sec:intro}

Multimodal Large Language Models (MLLMs) have achieved rapid progress in unified multimodal modeling and general-purpose reasoning~\citep{chen2024internvl25,wang2024qwen2vl,google2025gemini25}. In parallel, recent long-video benchmarks have extended evaluation from short clips to movies and other long-form videos~\citep{fu2024videomme,wu2024longvideobench,zhou2024mlvu}, while video LLM architectures have explored learnable long-term memory~\citep{song2024moviechat,he2024malmm} and streaming-efficient computation~\citep{zhang2024flashvstream}. Despite this progress, existing benchmarks still largely rely on synthetic or web-sourced videos composed of isolated sessions, where scenes change frequently and different clips rarely share persistent environments or cross-day dependencies.

Real-world egocentric video presents a fundamentally different challenge. Daily-life recordings are highly redundant yet temporally uneven: long periods may contain repeated routines or stable environments, while decisive events occur sparsely and may be separated by days or weeks. Meanwhile, familiar scenes often undergo subtle changes over time, requiring models to distinguish recurring background patterns from task-relevant details. Such properties make month-level understanding less a problem of dense frame perception than one of long-term memory: models must selectively retain salient evidence, retrieve sparse episodes, and maintain consistent temporal and spatial representations across extended experience.

Existing egocentric datasets, such as Ego4D~\citep{grauman2022ego4d}, have demonstrated the value of first-person scene-and-subject interaction for video understanding. However, their per-subject temporal depth and cross-day continuity remain insufficient for systematically evaluating persistent memory over month-scale daily life. As a result, it remains unclear whether current MLLMs can move beyond short-term clip perception to support long-term spatiotemporal memory and reasoning.

To address this gap, we propose \textbf{EgoMonth}, a month-level egocentric video understanding benchmark for evaluating long-term spatiotemporal memory in real-world daily-life settings. EgoMonth consists of longitudinally collected first-person videos spanning 20 to 120 days across diverse participants, devices, and environments. Unlike prior long-video benchmarks that mainly evaluate isolated video sessions, EgoMonth emphasizes per-participant continuity and cross-day reasoning. It systematically assesses how MLLMs consolidate recurring behavioral patterns, retrieve sparse episodic details, and integrate temporally distant evidence into coherent temporal, spatial, and multi-evidence reasoning. Our contributions are as follows:

\begin{enumerate}[leftmargin=*,itemsep=2pt,topsep=2pt]
  \item \textbf{Dataset.} We construct the first high-quality, month-level, egocentric, real-world multimodal long-video benchmark, comprising over 300 hours of video and 1{,}443 human-annotated QA pairs, including cross-video questions whose answers may require evidence from multiple videos, together with corresponding metadata for future extension.
  \item \textbf{Task Taxonomy.} We design a cognitively grounded 14-task evaluation framework organized into three hierarchical cognitive levels: Schema Consolidation, Episodic Indexing, and Cascading Reasoning. This framework systematically probes long-term memory stability and logical integration, ranging from recurring behavioral pattern discovery to transient detail retrieval and multi-evidence spatiotemporal reasoning.
  \item \textbf{Model Evaluation.} We conduct a comprehensive evaluation of representative open-source and closed-source MLLMs on EgoMonth under a unified multiple-choice protocol, providing a systematic comparison of their capabilities and remaining gaps in month-level egocentric video understanding.
  \item \textbf{Diagnostic Findings.} Our results reveal that current MLLMs remain far from human-level long-term spatiotemporal memory, especially on tasks requiring precise temporal indexing, stable spatial grounding, and multi-step cross-video reasoning over days or weeks. These findings expose key bottlenecks such as temporal attention dilution and spatial grounding collapse, offering guidance for future video LLM architectures.
\end{enumerate}


\begin{figure}[t]
    \centering

    \includegraphics[width=\linewidth, height=0.28\textheight, keepaspectratio]{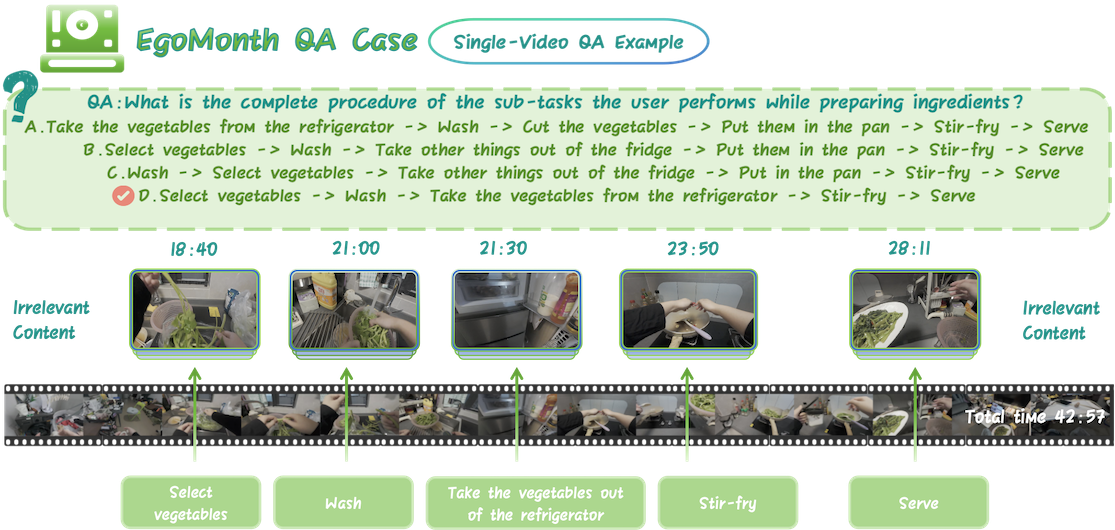}

    \vspace{2mm}

    \includegraphics[width=\linewidth, height=0.42\textheight, keepaspectratio]{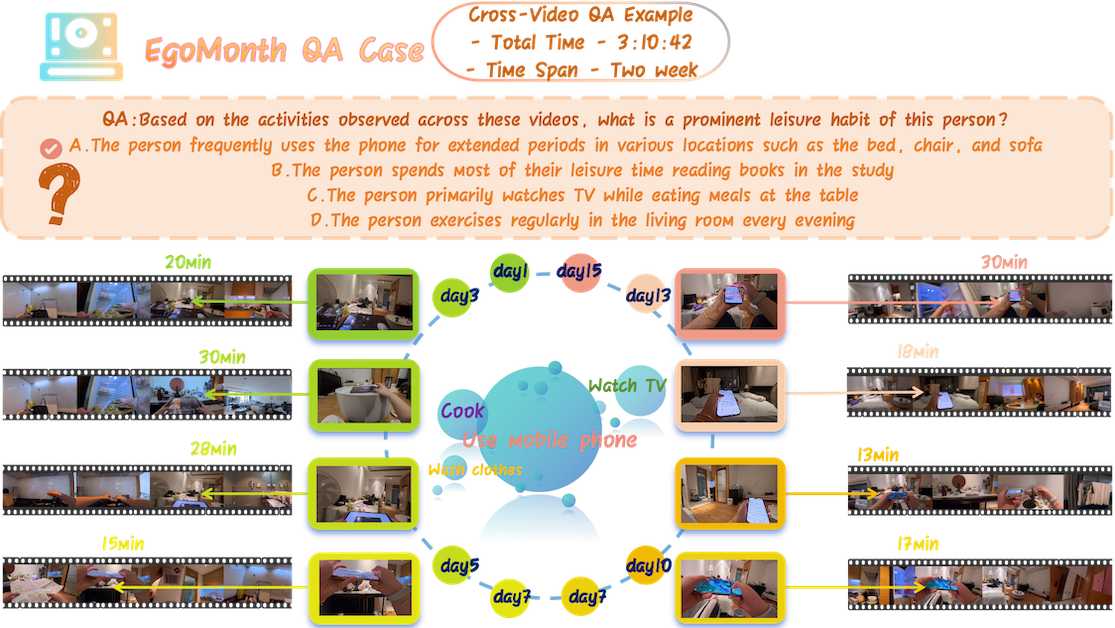}

    \caption{\textbf{Examples of EgoMonth QA design.} 
    The upper example shows a single-video QA case, while the lower example shows a cross-video QA case requiring evidence integration across multiple recordings.}
    \label{fig:qa_examples}
\end{figure}

\section{Related Work}
\label{sec:related}

\subsection{Benchmarks for Long Video Understanding}

Recent video understanding benchmarks have progressively expanded from short clips to long-form videos. Early benchmarks such as MMBench~\citep{liu2025mmbench}, SEED-Bench~\citep{li2024seedbench}, and MVBench~\citep{li2024mvbench} mainly focus on second-to-minute-level video understanding. Later benchmarks, including Video-MME~\citep{fu2024videomme}, LongVideoBench~\citep{wu2024longvideobench}, MLVU~\citep{zhou2024mlvu}, LVBench~\citep{wang2024lvbench}, HLV-1K~\citep{hlv1k2024}, and HourVideo~\citep{chandrasegaran2024hourvideo}, extend the temporal range to tens of minutes or even hours. These benchmarks have significantly advanced the evaluation of long-context video perception and reasoning. However, most of them rely on web videos, movies, or isolated video sessions, where clips do not share persistent environments, recurring routines, or cross-day dependencies.

Egocentric benchmarks provide a closer setting for evaluating embodied video understanding. Large-scale datasets such as Ego4D~\citep{grauman2022ego4d} and Ego-Exo4D~\citep{grauman2024egoexo4d} capture rich human-object and human-environment interactions, while EPIC-KITCHENS~\citep{damen2022epic}, EgoSchema~\citep{mangalam2023egoschema}, EgoThinker~\citep{pei2025egothinker}, EgoPlan-Bench~\citep{chen2024egoplan}, and MyEgo~\citep{xiao2026myego} further explore egocentric activity recognition, long-form QA, reasoning, planning, and personalized understanding. Nevertheless, these datasets primarily focus on short clips or single-session recordings, leaving cross-day and cross-week memory underexplored. In contrast, EgoMonth targets month-level per-participant continuity, enabling evaluation of longitudinal behaviors such as habits, routines, spatial familiarity changes, and cross-day causal reasoning.

\subsection{Lifelog and Multi-Day Egocentric Recording}

Lifelog research is closely related to EgoMonth because it studies long-term first-person experience. The NTCIR Lifelog task~\citep{gurrin2016ntcir} and the Lifelog Search Challenge~\citep{gurrin2022lsc} use multi-month camera streams for retrieval and concept detection. However, these works mainly rely on low-frame-rate image streams rather than continuous egocentric video, and their tasks are typically formulated as text-query-to-image retrieval instead of MLLM-based video understanding.

More recently, EgoLife~\citep{dong2025egolife} introduced a multi-day egocentric video benchmark collected from participants living together for one week. It represents an important step toward long-context egocentric QA, but differs from EgoMonth in temporal scale, data setting, and evaluation focus. EgoMonth extends the span from one week to 20--120 days, covers independent daily routines across diverse participants and environments, and emphasizes sparse event retrieval, long-term behavioral consolidation, and cross-day spatiotemporal reasoning. It therefore complements prior lifelog and egocentric datasets by testing whether MLLMs can maintain faithful memory over month-level real-world experience.

\begin{figure}[t]
    \centering
    \captionsetup[subfigure]{font=small, justification=centering, skip=3pt}

    \begin{subfigure}[b]{0.47\linewidth}
        \centering
        \raisebox{3mm}{%
            \includegraphics[width=\linewidth, height=0.22\textheight, keepaspectratio]{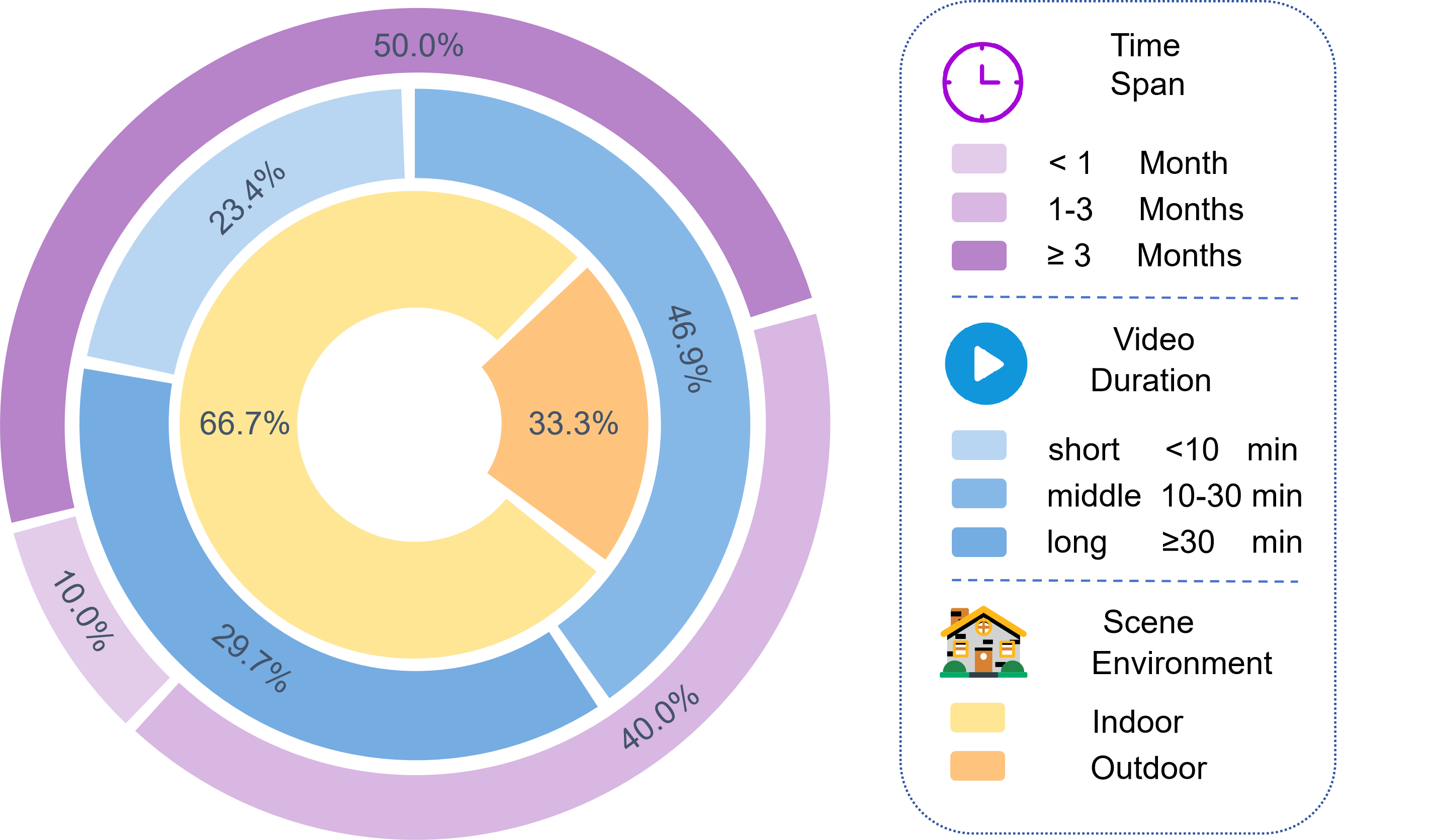}
        }
        \caption{Dataset Attribute Distribution}
        \label{fig:dataset_attribute}
    \end{subfigure}
    \hfill
    \begin{subfigure}[b]{0.50\linewidth}
        \centering
        \includegraphics[width=\linewidth, height=0.22\textheight, keepaspectratio]{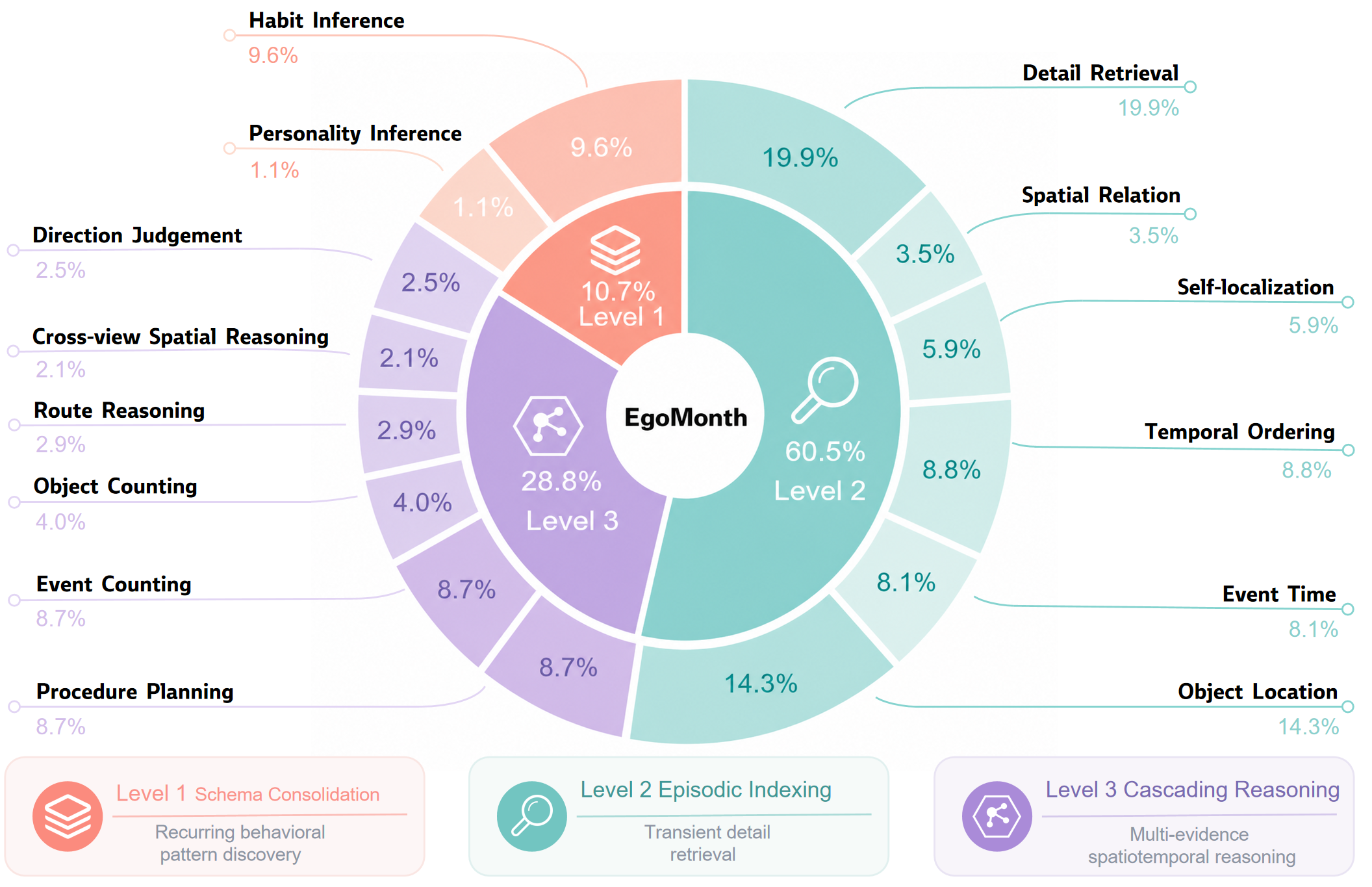}
        \caption{QA Pair Distribution}
        \label{fig:task_distribution}
    \end{subfigure}

    \vspace{-2mm}

    \caption{\textbf{Statistical analysis of the EgoMonth dataset.} 
    (a) Distributions of key dataset attributes, including video duration, collection time span, and scene environment.
    (b) QA-pair distribution across the 14 task types and three cognitive levels.}
    \label{fig:total_label}
\end{figure}

\section{EgoMonth Benchmark}
\label{sec:benchmark}

EgoMonth is constructed through video collection, quality screening, data anonymization, task framework design, human annotation, and cross-review.

\subsection{Dataset Construction and Anonymization}
\label{sec:construction}

We initially collect more than 400 hours of first-person daily-life video from 30 volunteers, with recording spans ranging from 20 to 120 days. 
To improve diversity and avoid overfitting to a single capture setup, participants are allowed to use different devices, including smartphones and action cameras, resulting in varied indoor and outdoor scenarios. 
We perform quality screening based on visual quality, temporal continuity, viewpoint stability, activity diversity, and relevance to long-term daily-life understanding. 
Low-quality segments, such as prolonged stationary recordings, ground-facing views, overly monotonous activities, or highly fragmented clips, are removed. 
After screening, we retain over 300 hours of high-quality video from 20 participants, with at least 1K resolution and a frame rate of at least 25 fps. 
To ensure privacy, we deploy an anonymization pipeline that integrates \textbf{Grounding DINO 1.5}~\citep{ren2024groundingdino15advance} for privacy-sensitive region detection and \textbf{SAM 2}~\citep{ravi2024sam} for precise instance segmentation and masking. 
Strong Gaussian blurring or masking is applied to detected regions, including faces of bystanders and non-participant individuals, personal identifiers such as license plates and home addresses, and private content on digital screens or physical documents. 
To reduce missed detections, automatically anonymized videos are further checked through manual quality inspection, and segments containing unresolved privacy risks are revised or excluded. 
This multi-stage approach mitigates privacy risks while preserving the essential spatiotemporal context for egocentric research.

\newcolumntype{C}{>{\centering\arraybackslash}X}


\begin{table}[t]
\caption{EgoMonth 14-task taxonomy organized by cognitive levels. The hierarchy reflects increasing demands on long-term memory, from redundant behavioral schema consolidation to sparse episodic indexing and multi-evidence spatiotemporal reasoning.}
\label{tab:task_taxonomy}
\centering
\footnotesize
\renewcommand{\arraystretch}{1.15} 
\setlength{\tabcolsep}{6pt}
\begin{tabularx}{\textwidth}{@{}c l X@{}}
\toprule
\textbf{Level} & \textbf{Task} & \textbf{Description} \\ \midrule

\multirow{2}{*}{\textbf{Level 1}} & 
Habit Inference & Inferring repetitive behavioral patterns over weeks \\
 & Personality Inference & Inferring long-term personality or character traits \\ \midrule

\multirow{6}{*}{\textbf{Level 2}} & 
Detail Retrieval & Fine-grained visual detail recall \\
 & Spatial Relation & Static spatial relationships between objects \\
 & Self-localization & Where the wearer is at a given moment \\
 & Temporal Ordering & Chronological ordering of events \\
 & Event Time & Temporal localization of specific events \\
 & Object Location & Where an object was last seen \\ \midrule

\multirow{6}{*}{\textbf{Level 3}} & 
Procedure Planning & Understanding multi-step procedures \\
 & Event Counting & Counting occurrences across days or weeks \\
 & Object Counting & Counting instances across the full video \\
 & Route Reasoning & Reconstructing spatial trajectories \\
 & Cross-view Spatial Reasoning & Inference across different viewpoints/days \\
 & Direction Judgement & Orientation perception from egocentric view \\ \bottomrule
\end{tabularx}
\end{table}

\subsection{Task Design Framework}
\label{sec:tasks}

Our task taxonomy follows a progressive cognitive hierarchy, moving from redundant long-term pattern consolidation, to sparse episodic indexing, and finally to multi-evidence reasoning. 
This design enables EgoMonth to evaluate whether models can go beyond perceiving individual clips to retain, retrieve, and integrate information over month-level egocentric experience. 
We organize the 14 tasks into three levels: \emph{Schema Consolidation}, \emph{Episodic Indexing}, and \emph{Cascading Reasoning} (Table~\ref{tab:task_taxonomy}). 
Figure~\ref{fig:task_distribution} shows the QA distribution, with detailed statistics provided in Appendix~\ref{app:participant} and Appendix~\ref{app:qa_dist}.

\begin{itemize}[leftmargin=*]

 \item \textbf{Level 1: Schema Consolidation.}
 Grounded in long-term memory formation~\citep{atkinson1968human}, this level evaluates whether models can infer stable behavioral schemas, such as habits or personality traits, from repeated cues across weeks. Since the same pattern may appear multiple times across clips, days, or environments, Level 1 tasks are relatively tolerant to local feature loss or imprecise temporal indexing.

 \item \textbf{Level 2: Episodic Indexing.}
 Drawing on episodic memory~\citep{tulving1972episodic}, this level requires models to locate specific, less redundant evidence within the long video stream, such as a particular object state, location, event time, spatial relation, or short temporal segment. Compared with Level 1, these tasks are more vulnerable to retrieval failure: if the model fails to access the correct moment, place, or object state, it may retrieve a visually similar but incorrect episode.

 \item \textbf{Level 3: Cascading Reasoning.}
 Building on the indexing ability required in Level 2, this level further requires models to retrieve, maintain, and compose multiple pieces of evidence across different times, locations, viewpoints, or action sequences. Inspired by working memory and cognitive load theories~\citep{baddeley2020working,sweller1994cognitive}, Level 3 tasks involve dependency-sensitive reasoning over multiple memory units, where a missing cue, inaccurate count, incorrect temporal index, or faulty spatial relation may propagate into a cascading failure of the final answer.

\end{itemize}

\subsection{Human Annotation and Cross-review}
\label{sec:annotation}

We construct 1{,}443 human-crafted QA pairs, each with four answer options. 
Annotators first browse each participant's long-term recordings to identify key events, scene transitions, object states, and temporal dependencies, which are then used to design questions requiring long-term retrieval or cross-video integration. 
To ensure answer reliability, all questions follow an unambiguous-answer principle and are verified through a cross-review process, where groups of three annotators check the correctness of answers and the plausibility of distractors. 
Unlike template-based or LLM-generated annotation, our questions are written in diverse free-form styles to better reflect natural long-term memory queries in real-world egocentric settings. 
The supporting evidence for each QA pair may come from either a single video or multiple videos across different days or recording sessions. Figure~\ref{fig:qa_examples} illustrates representative single-video and cross-video QA examples.

\subsection{Dataset Statistics}
\label{sec:statistics}

EgoMonth contains \textbf{738 video clips} from 20 participants, totaling \textbf{18{,}072 minutes ($\sim$301 hours)} of egocentric video. 
Recordings span \textbf{20 to 120 days} per participant, with an average clip duration of approximately 24.5 minutes. 
Based on this corpus, we construct \textbf{1{,}443} multiple-choice QA pairs.

We further analyze activity and environmental distributions to characterize dataset diversity. 
Fine-grained activities identified from sampled frames are grouped into six categories: \emph{study and work}, \emph{transportation}, \emph{food and dining}, \emph{shopping}, \emph{leisure and entertainment}, and \emph{household chores}. 
Environmental metadata collected during annotation covers both indoor and outdoor scenarios. 
Figure~\ref{fig:dataset_attribute} summarizes the dataset attribute distributions, including scene environment, video duration, and collection time span. 
The detailed activity category distribution is provided in Appendix~\ref{app:activities}. 
Together, these statistics show that EgoMonth captures diverse long-term daily-life experiences in realistic egocentric settings.

Table~\ref{tab:benchmark_comparison} compares EgoMonth with existing benchmarks across key dimensions. 
EgoMonth uniquely offers month-level per-participant continuity, cross-video reasoning, and high spatiotemporal consistency, filling a gap that no prior benchmark addresses.

\begin{table}[t]
\caption{Comparison of EgoMonth with existing long-video understanding benchmarks. 
EgoMonth is the only benchmark that supports cross-video reasoning with month-level per-participant continuity. 
Dashes (--) indicate that the corresponding statistic is not clearly reported or is not directly comparable across benchmarks. H, A, and LLM denote human annotation, automatic annotation, and LLM-assisted annotation, respectively.}
\label{tab:benchmark_comparison}
\centering
\footnotesize
\setlength{\tabcolsep}{2.2pt}
\renewcommand{\arraystretch}{1.05}
\begin{tabular}{@{}lcccccccc@{}}
\toprule
\textbf{Dimension} 
& \makecell{\textbf{Video-}\\\textbf{MME}} 
& \makecell{\textbf{LongVid}\\\textbf{Bench}} 
& \textbf{MLVU} 
& \makecell{\textbf{LV}\\\textbf{Bench}} 
& \makecell{\textbf{HLV}\\\textbf{-1K}} 
& \makecell{\textbf{Ego}\\\textbf{Schema}} 
& \makecell{\textbf{Ego}\\\textbf{Life}} 
& \makecell{\textbf{Ego}\\\textbf{Month}} \\
\midrule
Video source      
& Web     & Web     & Web     & Web     & Web     & Ego4D   & Ego     & Ego \\
Max video span    
& 60\,min & 60\,min & 2\,h    & 7\,h+   & $\sim$1\,h & 3\,min & 7\,d & 120\,d \\
Per-participant continuity 
& N/A     & N/A     & N/A     & N/A     & N/A     & N/A     & Week    & Month \\
Cross-video reasoning   
& No      & No      & No      & No      & No      & No      & Yes     & Yes \\
\#Videos / clips  
& 900     & 3{,}763 & 1{,}334 & 103     & 1{,}009 & 5{,}031 & -- & 738 \\
Total video duration   
& 254\,h  & --      & --      & 117\,h  & $\sim$925\,h & 250\,h+ & 300\,h & 301\,h \\
\#QA pairs / tasks
& 2{,}700 & 6{,}678 & 2{,}593 & 1{,}549 & 14{,}847 & 5{,}031 & 3{,}000 & 1{,}443 \\
Annotation        
& Human   & Human   & Human   & Human   & H+LLM   & A+H     & LLM+H   & Human \\
\#Task types      
& 4       & 17 & 9   & 6       & 4       & 1       & 5       & 14 \\
\bottomrule
\end{tabular}
\end{table}

\section{Experiments}
\label{sec:experiments}

\subsection{Experimental Setup}
\paragraph{Models.}
We evaluate twelve representative open-source and closed-source MLLMs:
\textbf{Chat-UniVi-V1.5} (7B, 256 frames)~\citep{jin2024chat},
\textbf{LLaVA-NeXT-Video} (7B, 64 frames)~\citep{zhang2024llavanextvideo},
\textbf{MiniCPM-V 4.5} (8B, 256 frames)~\citep{yu2025minicpm},
\textbf{Qwen2-VL} (7B, 256 frames)~\citep{wang2024qwen2vl},
\textbf{Qwen2.5-VL} (32B, 256 frames)~\citep{bai2025qwen25vl},
\textbf{Qwen3-VL} (8B, 256 frames)~\citep{bai2025qwen3vl},
\textbf{Qwen3-VL-30B-A3B} (30B, 256 frames)~\citep{bai2025qwen3vl},
\textbf{ShareGPT4Video} (8B, 64 frames)~\citep{chen2024sharegpt4video},
\textbf{ST-LLM} (256 frames)~\citep{liu2024stllm},
\textbf{VideoLLaMA3} (7B, 512 frames)~\citep{zhang2025videollama3},
\textbf{VITA-1.5} (7B, 16 frames)~\citep{fu2025vita},
and \textbf{Gemini 2.5 Pro} (1 fps)~\citep{google2025gemini25}.
For open-source models, we strictly follow their official configurations during evaluation. For closed-source models, we first adjust the video resolution according to the official video-input requirements and then evaluate the model through its API.

\paragraph{Protocol.} All QA pairs are in multiple-choice format with four options. We compare model outputs directly with ground truth without employing any LLM-as-judge. We report two metrics: \textbf{Avg} (macro-average of per-task accuracy) and \textbf{Acc} (micro-accuracy: total correct answers divided by total questions). The random-chance baseline is 25\% for four-option multiple-choice questions.

\paragraph{Human baseline.} Three trained annotators (not involved in QA creation) independently answer all 1{,}443 questions after watching the corresponding full videos. We report the macro-average human performance (Avg = \textbf{94.2\%}, Acc = \textbf{95.1\%}; inter-annotator Fleiss' $\kappa = 0.78$). Details are in Appendix~\ref{app:human}.

\subsection{Main Results}

\subsubsection{Performance Declines with Cognitive Complexity}
Results show a consistent performance decline along our three-level cognitive hierarchy. 
Models achieve the highest accuracy on Level 1 \emph{Schema Consolidation}, drop on Level 2 \emph{Episodic Indexing}, and perform worst on Level 3 \emph{Cascading Reasoning}. 
This trend confirms that increasing cognitive complexity poses a challenge for current MLLMs in month-level egocentric video understanding.

The gap between models and humans remains substantial. 
The best-performing model, \textbf{Gemini 2.5 Pro}, achieves 71.8\% macro-average accuracy, still 22.4 percentage points below the human baseline of 94.2\%. 
Although Gemini 2.5 Pro narrows the gap relative to the strongest open-source model, \textbf{Qwen2.5-VL} (32B) (58.0\%), the disparity remains especially pronounced in Level 3 tasks, indicating that current MLLMs still struggle with precise temporal indexing, stable spatial grounding, and multi-step cross-day reasoning over month-level egocentric videos.

\subsubsection{Model-wise Observations}
Among the evaluated models, \textbf{Gemini 2.5 Pro} achieves the best overall performance, obtaining the highest score across all task types. 
This suggests that stronger long-context multimodal reasoning can better exploit extended egocentric video evidence.

Among open-source models, \textbf{Qwen2.5-VL} (32B) achieves the best overall performance, with strong results on several Level 3 tasks, including \emph{Procedure Planning} (78.6\%), \emph{Object Counting} (50.9\%), and \emph{Route Reasoning} (52.4\%). 
However, performance does not scale uniformly with model size. 
For example, \textbf{MiniCPM-V 4.5} performs strongly on fine-grained Level 2 tasks such as \emph{Detail Retrieval} (66.5\%) and \emph{Temporal Ordering} (70.1\%), outperforming several larger models. 
By contrast, models primarily optimized for short-video understanding, such as \textbf{ShareGPT4Video} and \textbf{ST-LLM}, show consistently lower performance on EgoMonth.

\subsubsection{Task-specific Failure Patterns}
Task-level results reveal persistent weaknesses in temporal indexing and spatial reasoning. 
Tasks such as \emph{Cross-view Spatial Reasoning}, \emph{Self-localization}, and \emph{Direction Judgement} remain challenging for most models, with several results close to the 25\% random-chance level. 
Counting tasks are also difficult: even when models process hundreds of frames, they often fail to reliably count events or objects across days. 
Although \textbf{Gemini 2.5 Pro} substantially improves the upper bound of model performance, it still remains far below humans on several Level 3 tasks.

\begin{table}[t]
\caption{Per-task accuracy (\%) on EgoMonth. \textbf{Bold}: best score within each model group; human performance is shown as reference. Avg: macro-average across tasks. Acc: accuracy of all questions. Human: average of three annotators. Gray rows indicate tasks where at least one model falls below random chance (25\%).}
\label{tab:main_results}
\centering
\setlength{\tabcolsep}{1.2pt}
\scriptsize
\begin{tabularx}{\textwidth}{@{}l|ccccccccccc|c|c@{}}
\toprule
& \multicolumn{11}{c|}{\makecell{\textbf{Open-}\\\textbf{source}}} 
& \multicolumn{1}{c|}{\makecell{\textbf{Closed-}\\\textbf{source}}} 
& \multicolumn{1}{c}{\textbf{--}} \\
\cmidrule(lr){2-12}\cmidrule(lr){13-13}\cmidrule(l){14-14}
\textbf{Task} 
& \makecell{\textbf{Chat-}\\\textbf{UniVi}\\\textbf{V1.5}} 
& \makecell{\textbf{LLaVA-}\\\textbf{NeXT}\\\textbf{Video}} 
& \makecell{\textbf{MiniCPM}\\\textbf{V4.5}} 
& \makecell{\textbf{Qwen2}\\\textbf{VL}} 
& \makecell{\textbf{Qwen2.5}\\\textbf{VL}} 
& \makecell{\textbf{Qwen3}\\\textbf{VL}} 
& \makecell{\textbf{Qwen3}\\\textbf{VL-30B}\\\textbf{A3B}} 
& \makecell{\textbf{Share}\\\textbf{GPT}} 
& \makecell{\textbf{ST}\\\textbf{LLM}} 
& \makecell{\textbf{Video}\\\textbf{LLaMA3}} 
& \makecell{\textbf{VITA}\\\textbf{1.5}} 
& \makecell{\textbf{Gemini}\\\textbf{2.5 Pro}} 
& \textbf{Human} \\
\midrule
\# Frames     & 256 & 64 & 256 & 256 & 256 & 256 & 256 & 64 & 256 & 512 & 16 & 1fps & -- \\
\# Parameters & 7B  & 7B & 8B  & 7B  & 32B & 8B  & 30B & 8B & --  & 7B  & 7B & --   & -- \\
\midrule
\multicolumn{14}{l}{\textit{Level 1: Schema Consolidation}} \\
Habit Inference             
& 58.7 & 72.5 & \textbf{82.6} & 75.4 & 81.9 & 75.4 & 81.9 & 50.0 & 65.2 & 69.6 & 78.3 & \textbf{84.8} & 97.8 \\
Personality Inference       
& 50.0 & 62.5 & \textbf{68.8} & 56.2 & 62.5 & \textbf{68.8} & 37.5 & 50.0 & 56.2 & 50.0 & 56.2 & \textbf{81.3} & 93.8 \\
\midrule
\multicolumn{14}{l}{\textit{Level 2: Episodic Indexing}} \\
Detail Retrieval           
& 34.1 & 38.3 & \textbf{66.5} & 58.9 & 61.7 & 54.4 & 56.8 & 34.8 & 34.8 & 55.4 & 51.6 & \textbf{77.7} & 95.1 \\
Spatial Relation           
& 56.9 & 49.0 & \textbf{58.8} & 54.9 & 56.9 & 51.0 & 51.0 & 51.0 & 49.0 & 49.0 & 54.9 & \textbf{86.3} & 94.1 \\
Self-localization          
& 29.4 & 41.2 & \textbf{54.1} & 52.9 & 49.4 & 52.9 & 52.9 & 31.8 & 35.2 & 43.5 & 48.2 & \textbf{82.4} & 97.6 \\
Temporal Ordering          
& 40.9 & 45.7 & \textbf{70.1} & 59.8 & 64.6 & 55.9 & 62.2 & 36.2 & 40.1 & 62.2 & 57.5 & \textbf{75.6} & 95.3 \\
Event Time                 
& 44.4 & 46.1 & \textbf{57.3} & 42.7 & 47.9 & 41.9 & 50.4 & 45.3 & 42.7 & 41.9 & 49.6 & \textbf{60.7} & 95.7 \\
Object Location            
& 51.9 & 50.5 & 61.6 & 61.6 & \textbf{64.1} & 55.3 & 57.3 & 44.7 & 58.2 & 54.9 & 51.5 & \textbf{67.0} & 96.1 \\
\midrule
\multicolumn{14}{l}{\textit{Level 3: Cascading Reasoning}} \\
Procedure Planning             
& 46.0 & 29.4 & 68.2 & 68.2 & \textbf{78.6} & 71.4 & 70.6 & 30.2 & 46.0 & 60.3 & 66.7 & \textbf{81.7} & 94.4 \\
\rowcolor{gray!10}
Event Counting              
& 8.8  & 16.8 & 40.0 & 36.8 & \textbf{43.2} & 38.4 & 34.4 & 16.8 & 8.0  & 40.8 & 41.6 & \textbf{52.0} & 94.4 \\
Object Counting            
& 33.3 & 29.8 & 38.6 & \textbf{52.6} & 50.9 & 31.6 & 47.4 & 29.8 & 31.6 & 49.1 & 40.4 & \textbf{73.7} & 94.7 \\
\rowcolor{gray!10}
Route Reasoning              
& 35.7 & 23.8 & 35.7 & \textbf{52.4} & \textbf{52.4} & 28.6 & 45.2 & 30.9 & 28.6 & 45.2 & 40.5 & \textbf{64.3} & 90.5 \\
\rowcolor{gray!10}
Cross-view Spatial Reasoning 
& 23.3 & 30.0 & 43.3 & 40.0 & \textbf{50.0} & 46.7 & 46.7 & 26.7 & 20.0 & 40.0 & 33.3 & \textbf{60.0} & 93.3 \\
Direction Judgement            
& 38.9 & 33.3 & 38.9 & \textbf{50.0} & 47.2 & 47.2 & 47.2 & 41.7 & 27.8 & 41.7 & 47.2 & \textbf{58.3} & 86.1 \\
\midrule
\textbf{Avg}  
& 39.5 & 40.6 & 56.0 & 54.5 & \textbf{58.0} & 51.4 & 53.0 & 37.1 & 38.8 & 50.3 & 51.3 & \textbf{71.8} & \textbf{94.2} \\
\textbf{Acc}  
& 39.9 & 41.7 & 60.6 & 57.0 & \textbf{60.8} & 53.7 & 56.7 & 36.9 & 40.8 & 53.1 & 53.6 & \textbf{72.6} & \textbf{95.1} \\
\bottomrule
\end{tabularx}
\end{table}

\subsection{Analysis}

\subsubsection{Simply Increasing Frame Density Does Not Guarantee More Accurate Memory}
The results show that the number of input frames does not directly determine the accuracy of long-term memory. 
For example, \textbf{VITA-1.5} uses only 16 input frames but achieves 41.6\% on \emph{Event Counting}, outperforming several open-source models with substantially larger frame budgets, such as \textbf{Qwen2-VL} (7B) (36.8\%, 256 frames). 
By contrast, \textbf{Gemini 2.5 Pro} achieves the strongest overall performance with a 1 fps input rate. 
These results suggest that denser visual input can be useful, but only when the model can effectively select, align, and reason over the relevant evidence.

We identify two reasons. 
First, month-level egocentric videos contain substantial temporal redundancy, including repeated backgrounds and  routine actions. 
When more frames are introduced without effective evidence selection, decisive moments may be diluted by redundant visual tokens, making it difficult for the model to attend to the key frames relevant to the query. 
Second, dense sampling increases the demand for cross-frame correspondence modeling: the same object, action, or event may appear across adjacent frames or clips, while current models often fail to correctly align and map these observations into coherent temporal states. 
As a result, repeated views of the same event may be misinterpreted as independent occurrences, or temporally related evidence may fail to be integrated. 
Therefore, effective long-term video understanding requires models to better exploit additional frames through event-level temporal indexing, query-aware attention, and robust cross-frame correspondence, rather than relying on frame-level accumulation alone.

\subsubsection{Larger Scale Helps Evidence Integration, but Accurate Indexing Remains Critical}
Models with larger parameter scales often show advantages on tasks that require evidence integration and complex reasoning. 
Among open-source models, larger models such as \textbf{Qwen2.5-VL} (32B) perform relatively well on several Level 3 tasks, where models must maintain intermediate states, combine information across time or space, and reason over multi-step dependencies. 

However, increased parameter scale does not automatically guarantee accurate access to the right evidence. 
Many Level 2 tasks depend less on broad reasoning ability and more on precisely indexing a specific moment, object state, location, or temporal segment from a long video stream. 
If the model retrieves a visually similar but incorrect episode, larger parameter scale may still lead to a confident but wrong answer. 
Therefore, future progress requires not only scaling model parameters for complex reasoning, but also more reliable mechanisms for temporal localization, visual grounding, selective retrieval, and structured memory organization.

\subsubsection{More Structured Spatiotemporal Representations Are Needed for Long-Term Memory}
A central failure mode exposed by EgoMonth is the lack of sufficiently structured spatiotemporal representations. 
Month-level egocentric videos require models to maintain stable temporal indices and spatial relations over long, redundant, and visually repetitive recordings. 
Although current MLLMs may capture local events, objects, and scenes, they often fail to organize these observations into persistent temporal and spatial structures. 
As a result, models may retrieve a semantically related event but assign it to the wrong day, order, or occurrence, leading to errors in tasks such as \emph{Event Time}, \emph{Temporal Ordering}, \emph{Object Location}, and \emph{Event Counting}. 

Spatial reasoning shows a similar limitation. 
Egocentric videos provide partial and viewpoint-dependent observations, so solving \emph{Self-localization}, \emph{Route Reasoning}, \emph{Direction Judgement}, and \emph{Cross-view Spatial Reasoning} requires models to connect places, directions, routes, and viewpoints across days, rather than merely recognizing local landmarks. 
When such temporal or spatial relations are not represented in a stable structure, Level 3 reasoning chains can quickly collapse, producing cascading failures even when individual visual cues are recognizable. 
Future video MLLMs should therefore incorporate more explicit spatiotemporal structuring mechanisms, such as event-level temporal indices, persistent object-state tracking, and map-like spatial representations of rooms, routes, and landmarks, together with intermediate-state verification for complex multi-step reasoning.

\section{Limitations}
\label{sec:limitations}
Although EgoMonth establishes a month-level egocentric benchmark for long-term spatiotemporal memory, several directions remain for future expansion. 
First, the current version is built from 20 participants, which is sufficient for constructing a controlled benchmark, but a larger and more diverse participant pool would further improve demographic coverage and support more fine-grained subgroup analysis. 
Second, EgoMonth focuses on month-level to multi-month daily-life recordings. 
Extending the temporal span to seasonal or year-long recordings would allow future benchmarks to evaluate even longer-term memory, behavioral changes, and life-scale temporal reasoning.

\section{Broader Impact and Ethics} 
\label{sec:ethics} 
EgoMonth involves month-level first-person video and therefore raises privacy and misuse concerns. All participants provided informed consent for research use. Privacy-sensitive content, including bystander faces and personal identifiers, is anonymized before release. To reduce misuse risks, the dataset is distributed under a research-only license with tiered access control: metadata and QA pairs can be released for research purposes, while raw videos require additional approval and are stored on secured servers. Redistribution and commercial use are prohibited. Despite these risks, EgoMonth may support beneficial applications such as long-term personal AI agents, assistive technologies, cognitive support systems, and elderly-care tools that require persistent multimodal context over extended periods. We provide the consent-form template in Appendix~\ref{app:consent_form}.

\section{Conclusion}
\label{sec:conclusion}

We introduce EgoMonth, a month-level egocentric video understanding benchmark for evaluating long-term spatiotemporal memory. EgoMonth comprises over 300 hours of first-person daily-life recordings from 20 participants spanning 20 to 120 days, together with 1{,}443 human-crafted QA pairs across 14 tasks and three cognitive levels: Schema Consolidation, Episodic Indexing, and Cascading Reasoning. The QA pairs include cross-video questions whose answers may require evidence from multiple videos. Experiments on representative open-source and closed-source MLLMs show that current models remain far from human-level performance: the best-performing model, Gemini 2.5 Pro, achieves 71.8\% macro-average accuracy, compared with 94.2\% for humans. Our analysis further shows that accurate long-term memory depends not only on frame density or parameter scale, but also on accurate evidence indexing and more structured spatiotemporal representations. EgoMonth thus provides a challenging testbed and diagnostic framework for developing future video MLLMs with faithful long-term spatiotemporal memory.

\bibliographystyle{plainnat}
\bibliography{references}

\newpage
\appendix

\section{Per-Participant Dataset Statistics}
\label{app:participant}

Table~\ref{tab:participant_stats} provides detailed per-participant statistics. We initially recruit 30 volunteers; 10 are excluded due to participant attrition or insufficient recording quality, leaving 20 participants in the final dataset.

\begin{table}[h]
\caption{Per-participant statistics of the EgoMonth dataset. ``\#Clips'' denotes the number of video segments; ``Duration'' is total video length in minutes; ``Span'' is the temporal span from first to last recording day; ``Short'' counts clips $\leq$10 min.}
\label{tab:participant_stats}
\centering
\small
\begin{tabular}{@{}clrrrr@{}}
\toprule
\textbf{ID} & \textbf{Pseudonym} & \textbf{\#Clips} & \textbf{Duration (min)} & \textbf{Span (days)} & \textbf{Short ($\leq$10\,min)} \\
\midrule
1  & Participant-01  & 84  & 1{,}552 & 120 & 7 \\
2  & Participant-02  & 12  & 226     & 40  & 3 \\
3  & Participant-03  & 45  & 1{,}044 & 70  & 5 \\
4  & Participant-04  & 17  & 491     & 40  & 1 \\
5  & Participant-05  & 19  & 490     & 40  & 5 \\
9  & Participant-09  & 30  & 1{,}323 & 70  & 3 \\
10 & Participant-10  & 14  & 418     & 40  & 1 \\
11 & Participant-11  & 54  & 894     & 90  & 19 \\
12 & Participant-12  & 104 & 917     & 120 & 53 \\
13 & Participant-13  & 85  & 1{,}822 & 120 & 23 \\
14 & Participant-14  & 5   & 182     & 30  & 0 \\
15 & Participant-15  & 21  & 708     & 40  & 3 \\
16 & Participant-16  & 27  & 530     & 40  & 13 \\
17 & Participant-17  & 53  & 2{,}727 & 90  & 0 \\
18 & Participant-18  & 4   & 104     & 20  & 0 \\
19 & Participant-19  & 31  & 864     & 70  & 4 \\
20 & Participant-20  & 55  & 1{,}839 & 90  & 6 \\
21 & Participant-21  & 51  & 914     & 90  & 12 \\
23 & Participant-23  & 50  & 830     & 90  & 14 \\
24 & Participant-24  & 8   & 197     & 20  & 0 \\
\midrule
\multicolumn{2}{l}{\textbf{Total}} & \textbf{738} & \textbf{18{,}072} & -- & \textbf{172} \\
\bottomrule
\end{tabular}
\end{table}

\begin{figure}[t]
    \centering
    \includegraphics[width=0.65\linewidth, keepaspectratio]{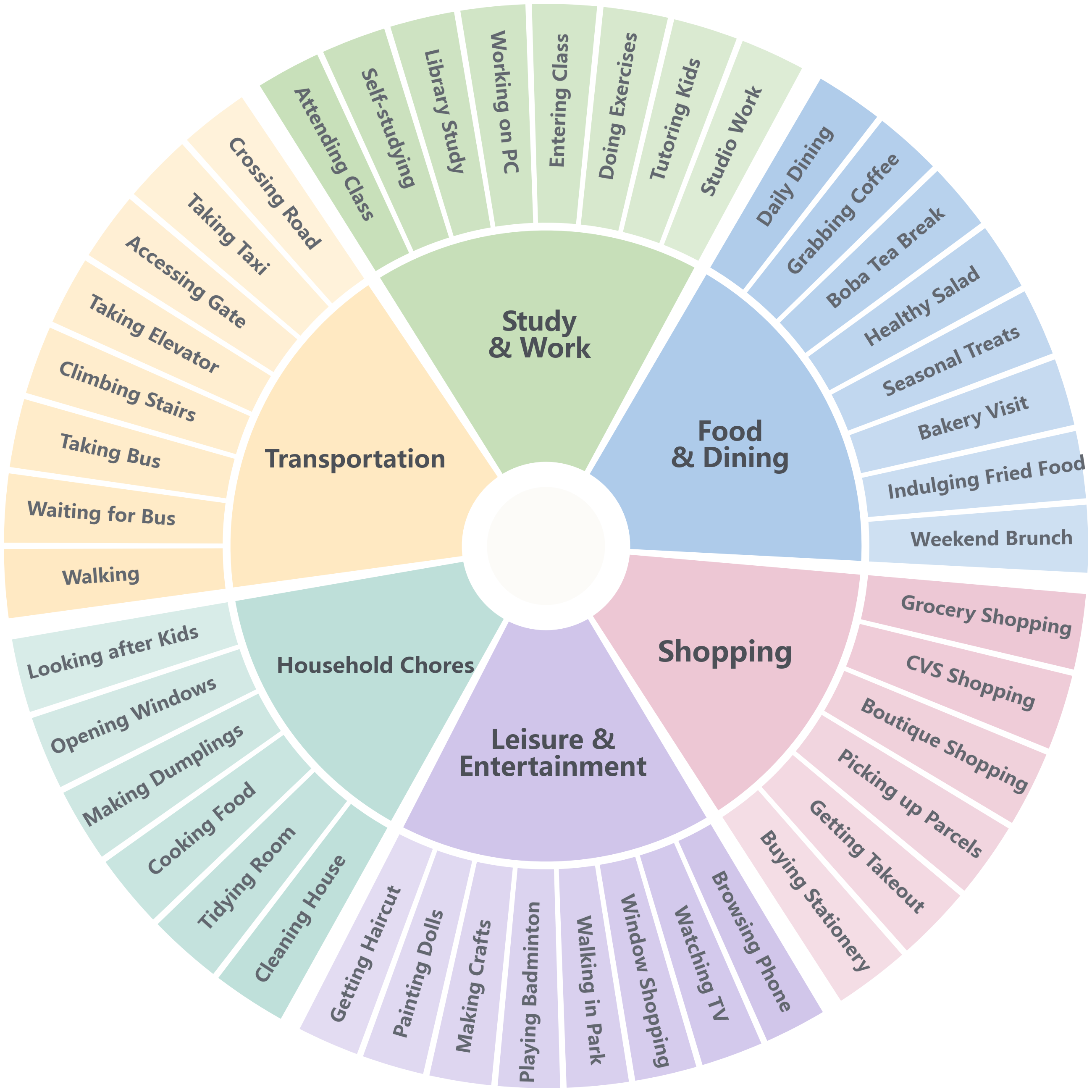}
    \caption{\textbf{Activity category distribution.} 
    EgoMonth covers six daily-life activity domains and 30 fine-grained activity types, providing broad semantic coverage of real-world egocentric daily life.}
    \label{fig:activity_categories_appendix}
\end{figure}
\section{Activity Categories}
\label{app:activities}

We organize the observed activities into six broad categories reflecting the diversity of daily-life scenarios captured in EgoMonth, as shown in Figure~\ref{fig:activity_categories_appendix}. 
Study and work includes attending class, self-study, studying in libraries, computer work, entering classrooms, doing exercises, tutoring children, and recording-studio work. 
Transportation includes walking, waiting for buses, riding buses, going up or down stairs, taking elevators, entering or leaving residential compounds, taking taxis, and crossing streets. 
Food and dining includes eating meals and snacks as well as purchasing drinks, bread, and fried food. 
Shopping includes supermarket and convenience-store visits, boutique browsing, and picking up deliveries or takeout. 
Leisure and entertainment includes phone use, watching TV, strolling, walking in parks, playing badminton, doing handicrafts, painting plaster figurines, and getting haircuts. 
Household chores includes cleaning, tidying rooms, cooking, making dumplings, opening windows, and childcare.

\section{Annotation Guidelines}
\label{app:annotation}

\subsection{Annotator Training}

All annotators undergo a two-phase training process. In the orientation phase, they study the 14-task taxonomy, review representative QA examples for each task type, and complete a calibration exercise on a held-out sample. This phase ensures that annotators understand the distinctions among \emph{Schema Consolidation}, \emph{Episodic Indexing}, and \emph{Cascading Reasoning}, as well as the expected evidence requirements for each task category. In the supervised-practice phase, each annotator completes one full participant-level sample under supervision. Project leads then review the resulting QA pairs, provide detailed feedback on question clarity, evidence grounding, answer uniqueness, and distractor quality, and certify the annotator before independent work begins.

\subsection{Pre-annotation Browsing and Evidence Marking}

Before writing questions, annotators perform a preliminary browsing pass over each participant's long-term recordings. During this stage, they mark key events, scene transitions, recurring routines, object states, locations, and temporal dependencies that may support long-term memory questions. These observations serve as evidence anchors for constructing questions that require retrieval from month-level egocentric experience rather than superficial recognition from isolated clips. In particular, annotators are encouraged to identify information that is temporally distant, visually similar across days, or dependent on the relationship between multiple events, since such cases better reflect the benchmark's goal of evaluating long-term spatiotemporal memory.

\subsection{Question Design Principles}

Question writing follows six principles. First, every question must have a clearly identifiable correct answer derivable from the video content alone, without relying on external knowledge or subjective interpretation. Second, questions should preferentially require information from temporally distant segments; single-clip-answerable questions are allowed only when they probe localized episodic evidence, such as \emph{Detail Retrieval}, \emph{Self-Localization}, or \emph{Event Time}. Third, questions should be grounded in explicitly observed evidence, including events, object states, locations, routines, or cross-segment dependencies identified during the browsing stage. Fourth, distractors must remain plausible by referring to entities, locations, actions, or events that genuinely appear in the video but belong to different temporal or spatial contexts. Fifth, each participant-level sample should include at least one question from each of the 14 task types whenever the corresponding evidence exists in the recording. Sixth, annotators are encouraged to write in a natural and diverse style rather than relying on fixed templates.

Unlike template-based or LLM-generated annotation, all QA pairs in EgoMonth are human-crafted. Annotators are encouraged to formulate questions in a free-form and conversational manner, including temporally grounded and personally contextualized queries, while still ensuring that each answer is unambiguous and verifiable from the video.

\subsection{Distractor Categories}

We classify distractors into four common types: temporal confusion, where the entity or event is correct but the time, day, or occurrence is wrong; spatial confusion, where the activity or object is correct but the location is wrong; entity substitution, where a similar-looking object, person, or scene is used in place of the target; and quantity perturbation, where counting questions use numerically close but incorrect options. These distractors are designed to be challenging but fair: they must be plausible with respect to the participant's video history, while remaining clearly incorrect under careful evidence verification.

\subsection{Quality Assurance and Cross-review}

To ensure annotation reliability, EgoMonth adopts a multi-stage quality assurance process. After initial question writing, QA pairs are reviewed by groups of three annotators. Reviewers independently verify whether the correct answer is fully supported by the video, whether the question wording is unambiguous, and whether all distractors are plausible but incorrect. They also check for common failure cases, including multiple plausible answers, insufficient visual evidence, overly generic questions, answer leakage from wording, and distractors that are either trivially wrong or unsupported by the video.

Disagreements are resolved through discussion among the reviewers and, when necessary, by re-checking the corresponding video segments. QA pairs that cannot be grounded in clear video evidence are revised or removed. This cross-review mechanism helps ensure both logical consistency and answer reliability across the final benchmark.

\section{Task-Level QA Distribution}
\label{app:qa_dist}

Table~\ref{tab:qa_distribution} shows the task-level distribution of the final benchmark set.

\begin{table}[h]
\caption{Task-level QA distribution of the final EgoMonth benchmark set. Percentages are computed from the 1,443 QA pairs and visualized in Figure~\ref{fig:task_distribution}.}
\label{tab:qa_distribution}
\centering
\small
\begin{tabular}{@{}lrr|lrr@{}}
\toprule
\textbf{Task} & \textbf{\#QA} & \textbf{\%} & \textbf{Task} & \textbf{\#QA} & \textbf{\%} \\
\midrule
Detail Retrieval              & 287 & 19.9 & Event Counting               & 125 & 8.7  \\
Object Location               & 206 & 14.3 & Event Time                   & 117 & 8.1  \\
Habit Inference               & 138 & 9.6  & Self-localization            & 85  & 5.9  \\
Temporal Ordering             & 127 & 8.8  & Object Counting              & 57  & 4.0  \\
Procedure Planning            & 126 & 8.7  & Spatial Relation             & 51  & 3.5  \\
Route Reasoning               & 42  & 2.9  & Direction Judgement          & 36  & 2.5  \\
Cross-view Spatial Reasoning  & 30  & 2.1  & Personality Inference        & 16  & 1.1  \\
\midrule
\multicolumn{6}{l}{\textbf{Total: 1,443 QA pairs (100.0\%)}} \\
\bottomrule
\end{tabular}
\end{table}

\section{Expanded Cross-Benchmark Comparison}
\label{app:expanded_comparison}

Table~\ref{tab:expanded_crossbench} provides an expanded comparison with additional benchmark metrics where available.

\begin{table}[t]
\caption{Expanded cross-benchmark comparison. Values represent reported accuracy (\%) collected from benchmark papers, technical reports, or official model cards. EgoMonth reports Avg. Dashes indicate unavailable, version-mismatched, split-mismatched, or source-mismatched results.}
\label{tab:expanded_crossbench}
\centering
\setlength{\tabcolsep}{2.0pt}
\scriptsize
\begin{tabular}{@{}lccccccc@{}}
\toprule
\textbf{Model} & \makecell{\textbf{HLV-}\\\textbf{1K}} & \makecell{\textbf{Video-}\\\textbf{MME}} & \makecell{\textbf{LV-}\\\textbf{Bench}} & \makecell{\textbf{MLVU}} & \makecell{\textbf{MMVU}} & \makecell{\textbf{LongVid}\\\textbf{Bench}} & \makecell{\textbf{Ego-}\\\textbf{Month}} \\
\midrule
Chat-UniVi-V1.5    & --              & 41.2             & --              & --              & --              & --              & 39.5 \\
ST-LLM             & --              & 38.6             & --              & --              & --              & --              & 38.8 \\
Qwen2-VL           & 62.57           & 63.3              & --              & --              & --              & --              & 54.5 \\
Qwen3-VL           & --              & 71.4             & 58.0            & 78.1            & 58.7            & --              & 51.4 \\
VideoLLaMA3        & --              & 66.2             & --              & 73.0            & 47.2            & --              & 50.3 \\
VITA-1.5           & --              & 56.1             & --              & --              & --              & --              & 51.3 \\
MiniCPM-V 4.5      & --              & 67.9             & 50.4            & 75.1            & --              & --              & 56.0 \\
ShareGPT4Video     & --              & 39.9             & --              & 34.2            & --              & 41.8            & 37.1 \\
LLaVA-NeXT-Video   & --              & --               & --              & --              & --              & 43.5            & 40.6 \\
Qwen2.5-VL-32B     & --              & 70.5             & 49.0            & --              & --              & --              & 58.0 \\
Qwen3-VL-30B-A3B   & --              & 74.5             & 62.5            & 81.3            & 59.8            & --              & 53.0 \\
Gemini 2.5 Pro     & --              & 84.8             & 69.0            & 81.2            & 72.2            & --              & 71.8 \\
\bottomrule
\end{tabular}
\end{table}

\section{Dataset Card}
\label{app:datacard}
EgoMonth version 1.0 contains 738 video clips totaling 301 hours together with 1{,}443 QA pairs. We recruited 30 participants and retained final data from 20 after quality screening and attrition; the participant pool spans ages 18--65 and includes participants from diverse occupations. Recordings cover 20--120 days per participant and were collected using smartphones and action cameras such as GoPro, Insta360, and DJI devices. Videos have at least 1K resolution and a frame rate of at least 25 fps. All QA pairs are written in English and are fully human-crafted without LLM-assisted generation. The dataset is distributed under a research-only license, with raw video requiring additional access approval. Data are hosted on secured institutional servers with access logging, and access requests are reviewed periodically as part of a planned expansion toward a larger participant pool. All participants provided written informed consent, bystander faces were blurred, and redistribution is prohibited.

\section{Human Baseline Details}
\label{app:human}

Three trained annotators who were not involved in the original QA creation independently answered the full set of 1{,}443 questions after watching the corresponding month-long videos. Each annotator had unlimited time and could re-watch any portion of the video. The reported human baseline, with a macro average of 94.2\% and a micro accuracy of 95.1\%, is the mean across the three annotators; per-task human performance appears in Table~\ref{tab:main_results} of the main paper. Inter-annotator agreement across all questions, measured by Fleiss' $\kappa$, is 0.78, indicating substantial agreement.

\section{Compute Resources}
\label{app:compute}

Open-source model evaluations were conducted on NVIDIA RTX 4090 (48 GB) GPUs. 
Gemini 2.5 Pro was evaluated through its official API and is therefore not included in the reported local GPU-hour estimate. 
Each model was evaluated on the full EgoMonth benchmark in a single pass without ensembling. Approximate wall-clock times per model:

\begin{itemize}[leftmargin=*,itemsep=1pt]
  \item LLaVA-NeXT-Video (64 frames): $\sim$17 hours
  \item MiniCPM-V 4.5 (256 frames): $\sim$58 hours
  \item Qwen3-VL (256 frames): $\sim$51 hours
  \item ShareGPT4Video (64 frames): $\sim$17 hours
  \item VideoLLaMA3 (512 frames): $\sim$86 hours
  \item VITA-1.5 (16 frames): $\sim$8 hours
  \item Chat-UniVi-V1.5 (256 frames): $\sim$40 hours
  \item Qwen2.5-VL (256 frames): $\sim$191 hours
  \item Qwen2-VL (256 frames): $\sim$54 hours
  \item ST-LLM (256 frames): $\sim$58 hours
  \item Qwen3-VL-30B-A3B (256 frames): $\sim$182 hours
\end{itemize}

Total compute for all open-source evaluations: approximately 762 GPU-hours on NVIDIA RTX 4090 (48 GB) GPUs.


\section{Informed Consent Form for Research Volunteers}
\label{app:consent_form}
For participant privacy, this appendix provides a blank template of the informed consent form. 
All signed consent forms are securely retained by the research team and are not included in the paper.
\small

\noindent\textbf{Research Project Title:} EgoMonth: Real-World Dataset Collection

\noindent\textbf{Research Institution: } [Anonymized for review]

\subsubsection*{I. Research Background and Purpose}

This study aims to construct a large-scale first-person perspective dataset to advance research in computer vision, multimodal large language models (MLLMs), and video understanding. By collecting visual and environmental data from volunteers in real-world physical settings over extended periods, including daily behavioral patterns, lifestyle habits, and environmental variations, we seek to contribute to the development of more intelligent models and agent systems capable of understanding and interacting with human daily activities.

\subsubsection*{II. Data Collection Procedures and Participant Tasks}

\begin{enumerate}[leftmargin=*, itemsep=2pt, topsep=2pt]
    \item \textbf{Device Usage:} Participants will wear portable recording devices, such as GoPro cameras or similar mobile equipment. These devices will capture video and audio data during the study period.

    \item \textbf{Duration of Participation:} Data collection is expected to last approximately 30 days, with an estimated recording duration of around half an hour per day; longer-term participation is also welcome when participants are willing and available.

    \item \textbf{Recording Scenarios:} Data collection will take place in authentic daily-life settings, including but not limited to office work, dining, shopping, walking, and social interactions.

    \item \textbf{Participant Control:} During the data collection process, participants retain the right to deactivate the recording device at any time or request deletion of recordings from specific periods.
\end{enumerate}

\subsubsection*{III. Data Privacy and Protection}

We recognize that first-person perspective data may involve highly sensitive personal information. To ensure participant privacy and data security, the following protective measures will be implemented:

\begin{enumerate}[leftmargin=*, itemsep=2pt, topsep=2pt]
    \item \textbf{Data De-identification:} Prior to annotation or analysis, all collected data will undergo manual review to identify and remove potentially sensitive or personally identifiable information.

    \item \textbf{Audio Filtering:} Audio recordings will be muted or removed unless audio data is essential for the research objectives.

    \item \textbf{Anonymization:} Participants' real names, identification numbers, and other personally identifying information will not be associated with the dataset. Instead, anonymized participant IDs will be used.

    \item \textbf{Secure Storage:} Raw data will be stored on physically protected and encrypted laboratory servers, accessible only to authorized project team members. Any publicly released version of the dataset will have restricted access controls, and all video data will be used solely for scientific research purposes, with commercial use strictly prohibited.
\end{enumerate}

\subsubsection*{IV. Potential Risks and Participant Rights}

\paragraph{Participant Rights.}

\begin{enumerate}[leftmargin=*, itemsep=2pt, topsep=2pt]
    \item \textbf{Voluntary Participation:} Participation in this study is entirely voluntary.

    \item \textbf{Right to Withdraw:} Participants may discontinue their involvement at any stage of the data collection process without providing any reason. Withdrawal will not affect their eligibility for volunteer service hours or financial compensation.

    \item \textbf{Data Withdrawal:} Prior to official public release of the dataset, participants may contact the research team to request deletion of their data.
\end{enumerate}

\subsubsection*{V. Compensation and Remuneration}

Upon successful completion of the data collection task, participants will receive:

\begin{itemize}[leftmargin=*, itemsep=2pt, topsep=2pt]
    \item \textbf{Volunteer Compensation:} RMB 500 for 30 hours of participation.
\end{itemize}

\subsubsection*{VI. Signature of Consent}

\noindent\textbf{Volunteer Statement:} I have carefully read and fully understood the contents of this informed consent form. I have had the opportunity to ask questions regarding the study and have received satisfactory answers. I voluntarily agree to participate in this research project and consent to the data handling procedures described above.

\vspace{1.5em}

\noindent\textbf{Volunteer Signature:} \rule{0.45\textwidth}{0.4pt}

\vspace{1em}

\noindent\textbf{Date:} \rule{0.12\textwidth}{0.4pt} / \rule{0.12\textwidth}{0.4pt} / \rule{0.18\textwidth}{0.4pt}

\normalsize

\newpage
\section*{NeurIPS Paper Checklist}

\begin{enumerate}

\item {\bf Claims}
    \item[] Question: Do the main claims made in the abstract and introduction accurately reflect the paper's contributions and scope?
    \item[] Answer: \answerYes{}
    \item[] Justification: The abstract and introduction clearly state four contributions (dataset, task taxonomy, Model Evaluation, diagnostic findings) that are supported by the experimental results in Section~4.

\item {\bf Limitations}
    \item[] Question: Does the paper discuss the limitations of the work performed by the authors?
    \item[] Answer: \answerYes{}
    \item[] Justification: Section~5 discusses limitations including participant scale and temporal coverage.

\item {\bf Theory assumptions and proofs}
    \item[] Question: For each theoretical result, does the paper provide the full set of assumptions and a complete (and correct) proof?
    \item[] Answer: \answerNA{}
    \item[] Justification: This paper is a dataset and benchmark contribution and does not include theoretical results.

    \item {\bf Experimental result reproducibility}
    \item[] Question: Does the paper fully disclose all the information needed to reproduce the main experimental results of the paper to the extent that it affects the main claims and/or conclusions of the paper (regardless of whether the code and data are provided or not)?
    \item[] Answer: \answerYes{}
    \item[] Justification: Section~4.1 specifies all model names, parameter counts, frame counts, and evaluation protocols. Appendix~H provides compute resources.

\item {\bf Open access to data and code}
    \item[] Question: Does the paper provide open access to the data and code, with sufficient instructions to faithfully reproduce the main experimental results, as described in supplemental material?
    \item[] Answer: \answerYes{}
    \item[] Justification: We provide an anonymized reviewer-accessible package at submission time, including QA metadata, a small anonymized video sample for data-quality inspection, and evaluation code with instructions for reproducing the multiple-choice benchmark protocol.

\item {\bf Experimental setting/details}
    \item[] Question: Does the paper specify all the training and test details (e.g., data splits, hyperparameters, how they were chosen, type of optimizer) necessary to understand the results?
    \item[] Answer: \answerYes{}
    \item[] Justification: Section~4.1 details model configurations, frame sampling strategies, and evaluation metrics. This is a benchmark evaluation (no training involved).

\item {\bf Experiment statistical significance}
    \item[] Question: Does the paper report error bars suitably and correctly defined or other appropriate information about the statistical significance of the experiments?
    \item[] Answer: \answerNo{}
    \item[] Justification: Model evaluations are deterministic (greedy decoding, fixed frame sampling). We report inter-annotator agreement for the human baseline (Fleiss' $\kappa$ = 0.78 in Appendix~G).

\item {\bf Experiments compute resources}
    \item[] Question: For each experiment, does the paper provide sufficient information on the computer resources (type of compute workers, memory, time of execution) needed to reproduce the experiments?
    \item[] Answer: \answerYes{}
    \item[] Justification: Appendix~H reports the GPU type (NVIDIA RTX 4090, 48 GB), per-model wall-clock time, and total compute hours.

\item {\bf Code of ethics}
    \item[] Question: Does the research conducted in the paper conform, in every respect, with the NeurIPS Code of Ethics \url{https://neurips.cc/public/EthicsGuidelines}?
    \item[] Answer: \answerYes{}
    \item[] Justification: The research conforms with the NeurIPS Code of Ethics. Ethical considerations are discussed in detail in Section~6.

\item {\bf Broader impacts}
    \item[] Question: Does the paper discuss both potential positive societal impacts and negative societal impacts of the work performed?
    \item[] Answer: \answerYes{}
    \item[] Justification: Section~6 discusses privacy and misuse concerns, mitigation strategies for controlled dataset release, and potential beneficial applications such as assistive technologies and elderly-care tools.

\item {\bf Safeguards}
    \item[] Question: Does the paper describe safeguards that have been put in place for responsible release of data or models that have a high risk for misuse (e.g., pre-trained language models, image generators, or scraped datasets)?
    \item[] Answer: \answerYes{}
    \item[] Justification: Section~6 describes tiered access control, a research-only license, secured storage for raw videos, and prohibitions on redistribution and commercial use.

\item {\bf Licenses for existing assets}
    \item[] Question: Are the creators or original owners of assets (e.g., code, data, models), used in the paper, properly credited and are the license and terms of use explicitly mentioned and properly respected?
    \item[] Answer: \answerYes{}
    \item[] Justification: All evaluated models are cited with their original publications. The dataset is entirely original (no existing assets reused).

\item {\bf New assets}
    \item[] Question: Are new assets introduced in the paper well documented and is the documentation provided alongside the assets?
    \item[] Answer: \answerYes{}
    \item[] Justification: Appendix~F provides a dataset card following the datasheet template. Appendix~C documents annotation guidelines.

\item {\bf Crowdsourcing and research with human subjects}
    \item[] Question: For crowdsourcing experiments and research with human subjects, does the paper include the full text of instructions given to participants and screenshots, if applicable, as well as details about compensation (if any)?
    \item[] Answer: \answerYes{}
    \item[] Justification: Appendix~I provides the informed consent form and participant instructions, including device usage, participation duration, participant rights, data withdrawal, and compensation. Appendix~C provides annotation guidelines and training procedures for annotators.

\item {\bf Institutional review board (IRB) approvals or equivalent for research with human subjects}
    \item[] Question: Does the paper describe potential risks incurred by study participants, whether such risks were disclosed to the subjects, and whether Institutional Review Board (IRB) approvals (or an equivalent approval/review based on the requirements of your country or institution) were obtained?
    \item[] Answer: \answerYes{}
    \item[] Justification: EgoMonth involves human participants who voluntarily contribute month-level first-person recordings. Section~6 states that all participants provided informed consent for research use, and Appendix~I provides the consent-form template. The study describes potential privacy risks associated with first-person video and mitigates them through anonymization, secure storage, tiered access control, and restrictions on redistribution and commercial use.

\item {\bf Declaration of LLM usage}
    \item[] Question: Does the paper describe the usage of LLMs if it is an important, original, or non-standard component of the core methods in this research?
    \item[] Answer: \answerYes{}
    \item[] Justification: We used advanced MLLMs only as auxiliary tools for dataset-level statistical analysis, specifically to analyze sampled video frames and identify fine-grained activity types and recurring behavioral patterns for constructing activity distributions. LLMs or MLLMs were not used to generate QA pairs, ground-truth answers, distractors, annotations, or model-evaluation judgments. All QA pairs were human-crafted and cross-reviewed. Privacy anonymization was performed using computer vision models such as Grounding DINO 1.5 and SAM 2, which are not LLMs.

\end{enumerate}

\end{document}